\documentclass[10pt, a4paper, hidelinks]{article}

\usepackage[final]{lrec2026} 
\usepackage{graphicx}
\usepackage{amsmath} 
\usepackage{amssymb}
\usepackage{placeins}
\usepackage{booktabs}

\title{ 
Improving Neural Argumentative Stance Classification in Controversial Topics with Emotion-Lexicon Features  
}

\name{
\normalsize\bfseries
Mohammad Yeghaneh Abkenar\textsuperscript{1,3} \quad
Weixing Wang\textsuperscript{2} \\\bfseries
Manfred Stede\textsuperscript{3} \quad
Davide Picca\textsuperscript{4} \quad
Mark A. Finlayson\textsuperscript{5} \quad
Panagiotis Ioannidis\textsuperscript{6}
}
\address{
\textsuperscript{1}Innovations Department, Bundesdruckerei GmbH, Berlin, Germany \\
\textsuperscript{2}Hasso-Plattner-Institut, University of Potsdam, Germany \\
\textsuperscript{3}Department of Linguistics, University of Potsdam, Germany \\
\textsuperscript{4}University of Lausanne, Switzerland,  \textsuperscript{5}Knight Foundation School of Computing \& Information Sciences, \\ Florida International University, USA,
\textsuperscript{6}Pisquared, Germany \\ \\
yeghanehabkenar@uni-potsdam.de, 
weixing.wang@hpi.de \\
stede@uni-potsdam.de,
davide.picca@unil.ch, 
markaf@fiu.edu, 
pioannidis@pisquared.io
}

\abstract{
Argumentation mining comprises several subtasks, among which stance classification focuses on identifying the standpoint expressed in an argumentative text toward a specific target topic. While arguments---especially about controversial topics---often appeal to emotions, most prior work has not systematically incorporated explicit, fine-grained emotion analysis to improve performance on this task. In particular, prior research on stance classification has predominantly utilized non-argumentative texts and has been restricted to specific domains or topics, limiting generalizability. We work on five datasets from diverse domains encompassing a range of controversial topics and present an approach for expanding the Bias-Corrected NRC Emotion Lexicon using DistilBERT embeddings, which we feed into a Neural Argumentative Stance Classification model. Our method systematically expands the emotion lexicon through contextualized embeddings to identify emotionally charged terms not previously captured in the lexicon. Our expanded NRC lexicon (eNRC) improves over the baseline across all five datasets (up to +6.2 percentage points in $F_1$ score), outperforms the original NRC on four datasets (up to +3.0), and surpasses the LLM-based approach on nearly all corpora. We provide all resources---including eNRC, the adapted corpora, and model architecture---to enable other researchers to build upon our work. 
 \\ \newline  }

\begin{document}

\maketitleabstract

\section{Introduction}
Argumentation has long been central to human discourse. In ancient Greece (4th century BCE), Aristotle systematically analyzed rhetoric and persuasion. He developed modal logic theory which described three modes of persuasion: \textit{Pathos} (emotions and values), \textit{Ethos} (credibility and authority), and \textit{Logos} \citep[logical reasoning; ][]{schiappa2010keeping}. Meanwhile, in China (5th century BCE), Confucius developed theories of and approaches to ethical reasoning \citep{lu1998rhetoric}. Centuries later, Persian polymaths Al-Farabi (9th--10th century CE) and Avicenna (Ibn Sina) (10th--11th century CE) refined and expanded Aristotelian logic, creating comprehensive theories of dialectics, syllogistic reasoning, and rhetoric \citep{sep-arabic-islamic-language}. More recently, in the 20th century, Toulmin, in his famous book \textit{The Uses of Argument} (\citeyear{Toulmin1958-TOUTUO-6}), defined arguments as comprising six components—\textit{data} (premise), \textit{claim}, and \textit{warrant} being required, while \textit{backing}, \textit{qualifier} and \textit{ rebuttal} being optional.  Freeman's approach (\citeyear{freeman2011argument}) integrates the standard model with Toulmin's framework, using five elements: premise, conclusion, modality, rebuttal, and counter-rebuttal. Premise and conclusion (Claim) are the basic elements. For practical argumentation mining applications, this is typically simplified to premises and claims. For example, \textit{``Climate change threatens coastal cities''} (premise) supports \textit{``We must reduce carbon emissions''} (claim).

Related to argumentation, discussions about controversial topics have taken many forms throughout history: from Munazara (formal debate) in madrasas and the Athenian agora, to European parliamentary debates shaping European democracies, town hall meetings in early America, 20th century mass media debates, and today's online micro blogging forums like \textit{Reddit} and \textit{Quora} alongside televised presidential debates. Controversies serve a productive function when contributions successfully delineate proponent and opponent standpoints, facilitating the organization of discourse into identifiable argumentative clusters \citep{vecchi-etal-2021-towards}. Together, these forms demonstrate how argumentation about controversial topics remains vital to human communication and decision-making. 

The continuing interest in argumentation about controversial topics motivates our focus, but there are also significant practical applications. For example, robust stance classification can contribute to computational approaches to analyze public discourse at scale, identify majority and minority perspectives, distinguish genuine sentiment from amplified extremes, and detect misinformation campaigns that manipulate public opinion on controversial topics. Such capabilities are becoming increasingly valuable in a highly polarized media ecosystem.

\subsection{Contributions}

We focus here on the use and expansion of emotion lexicons for one of the less addressed subtasks: argumentative stance classification on controversial topics. Controversial issues involve fundamental value conflicts that evoke affective responses—emotional framing (e.g., fear, anger, hope, disgust) often reveals underlying stances more reliably than propositional content alone. Empirically, emotion lexicons enable knowledge transfer across controversial targets \citep{zhang-etal-2020-enhancing-cross}, improving stance detection on diverse contemporary issues \citep{hosseinia-etal-2020-stance}.

Our work makes two main contributions. First, we demonstrate a new method for expanding existing emotion lexicons through a clustering-based method using DistilBERT embeddings and similarity metrics. We apply this method to the Bias-Corrected version of the NRC lexicon \citep{zad2021hell} to produce the expanded NRC lexicon (eNRC). Second, we present a novel neural argumentative stance classification framework for leveraging the emotion lexicon to improve classifying stance in argumentative text on controversial topics. This framework involves a redefinition of the stance classification task for the argumentation mining corpora, achieved through extensive pre-processing of five well-known datasets and the integration of an emotion lexicon adapter. We release all code and resources to ensure full reproducibility.\footnote{For purposes of review, our expansion of the NRC lexicon \href{https://anonymous.4open.science/r/eNRC-5025/}{\textbf{eNRC}} may be found online, while the code for the experiments and model may be found at \href{https://anonymous.4open.science/r/ArgStanceNRC-47FB/}{\textbf{ArgStanceNRC}}. We will archive these objects in a permanent institutional repository if the paper is accepted for publication.}

\begin{figure*}[t]
\centering
\includegraphics[width=.60\textwidth ]{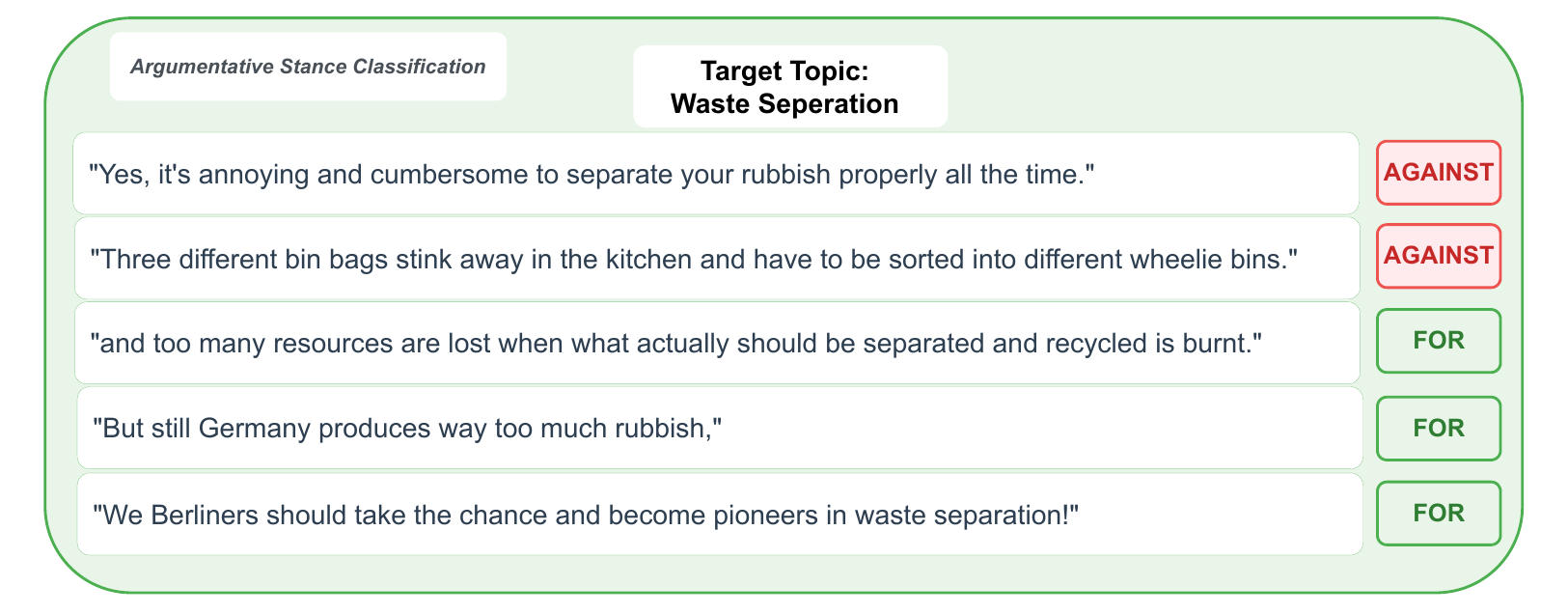}
\caption{An example with five segments from the Argumentative Microtext corpus (Part 1), showing stance labels (For or Against) toward a controversial target topic.}
\label{fig:diag.png}
\end{figure*}

\section{Background \& Related work}

\subsection{Background}
\label{sec:append-how-prod}

According to \citet{bentahar2010taxonomy}, there are three types of models of argumentation: \textit{monological} models, \textit{dialogical} models, and \textit{rhetorical} models. Monological models focus on the internal structure of arguments and relations among components (microstructure). Dialogical models focus on relationships between arguments while ignoring internal structures. Finally, rhetorical models consider rhetorical structure and patterns. Given that stance classification primarily concerns the internal structure of arguments and relationships among components, our work is situated primarily within work on monological models. All our corpora either have microstructure annotations or are adapted accordingly. Argumentation microstructures are closely tied to discourse analysis and frameworks such as RST \citep{mann1987rhetorical,hewett-etal-2019-utility}, which models discourse structure as a tree, PDTB \citep{prasad2017penn}, which focuses more on shallow structure, and SDRT \citep{asher2003logics}, which encompasses both logical and rhetorical structural elements. Despite this focus, argumentation inherently has dialogical aspects \citep{freeman2011argument}, making it suitable for analyzing controversial topics, a characteristic partially reflected in the structure of our corpora.

We organized our review of the background into three sections: \textit{argumentation mining}, \textit{emotion lexicons}, and \textit{stance classification}, since these are closest to the focus of this paper.

\paragraph{Argumentation Mining}

Argumentation mining as an topic of research has evolved rapidly in recent years with the widespread use of language models resulting in numerous workshops at major NLP conferences \citep{chistova-etal-2025-proceedings}. The term ``argument mining'' is an umbrella term for several subtasks, such as \textit{argument component type classification} (ACTC), \textit{argumentative relation classification} (ARC), \textit{argument structure identification}, \textit {argument stance classification}, and \textit{argument quality assessment}. However, despite the important connections among these subtasks, each is inherently complex and poses distinct challenges. Therefore, it is reasonable to address each of these subtasks separately. Moreover, as discussed in \citet{feger-etal-2025-limited}, the goal is to learn the task itself rather than merely memorizing the dataset. This paper focuses on one of the less-addressed subtasks, namely argumentative stance classification (identifying an author's standpoint on a topic), and evaluates the approach on five well-known corpora. 

\paragraph{Emotion Lexicons}

An emotion lexicon is a specialized linguistic resource that connects the emotional words in a language to predefined emotion categories such as Plutchik's eight basic emotions \citep{plutchik1965emotion}. Each word in the lexicon is linked to one or more emotion labels, or sometimes none at all \citep{mohammad-2023-best}. Prominent emotion lexicons include the General Inquirer \citep{stone1966general}, ANEW \citep{nielsen2011new}, the Pittsburgh Subjectivity Lexicon \citep{wilson2005opinionfinder}, the NRC Emotion Lexicon and it's updated versions \citeplanguageresource{mohammad2013nrc,mohammad2010emotions,zad2021hell}, the NRC Valence, Arousal, and Dominance (VAD) Lexicon \citep{mohammad2025nrc}, and SenticNet \citep{cambria2016senticnet}, all of which were created through manual annotation by experts or crowdsourcing. For our stance classification model, we use the eight emotion categories from the NRC Emotion Lexicon (\textit{joy}, \textit{trust}, \textit{fear}, \textit{surprise}, \textit{sadness}, \textit{disgust}, \textit{anger}, and \textit{anticipation}) as features and propose an embedding-based expansion (eNRC) that extends emotion labels to semantically similar words while preserving categorical structure. 

\paragraph{Stance Classification}
 One of the earliest works addressed detecting support or opposition to controversial legislation from Congressional floor debates. Interestingly, they framed this as document-level sentiment-polarity classification without using the term ``stance'' \citep{thomas-etal-2006-get}.
Stance can be defined as being for or against a defined target, such as a controversial topic, e.g., being \textit{for} or \textit{against} ``waste separation'' (a case illustrated in figure~\ref{fig:diag.png}), ``charge for plastic bags'', ``technology makes children even more creative'', ``multiculturalism'', or ``death penalty''.

Another relevant genre for stance classification is argumentative student essays responding to pro/con prompts. \citet{faulkner2014automated} estimated essay-level stance using on-topicness scoring via Wikipedia link similarity and stance-oriented dependency parsing to detect pro/con subtrees. The task of stance detection was later popularized and standardized by \citet{mohammad-etal-2016-semeval}, which focused on Twitter data. Stance classification (detection) is a well-established task, often studied separately from argument mining more broadly \citep{Kk2020StanceD}.
 In the next subsection on related work, we focus primarily on approaches that explicitly incorporate arguments into their stance models. The emphasis is on the intersection of argumentation mining, stance classification, and emotion analysis.






\subsection{Related Work}

There has been limited work on the intersection of argumentative stance classification, emotions, and controversial topics. For example, \citet{sobhani2015argumentation} found that using automatically-extracted arguments as features for stance classification yielded promising results. However, this work was also relying on TF-IDF features and predicted argument tags with a linear SVM classifier. They compared this to a TF-IDF-only linear SVM and a majority-class baseline. In contrast, we developed a sophisticated end-to-end neural framework for argumentative stance classification, evaluated across multiple domains. Although we did not use explicit argument features for the stance classification task, our approach implicitly captures argumentative and emotional cues through its context-aware neural architecture. Furthermore, \citet{bar2017stance} presented pivotal work in stance classification using the IBM dataset \citet{aharoni-etal-2014-benchmark}. However, their approach focused on a single corpus and employed sentiment rather than emotion features.
\citet{schaefer2019improving} utilized the Atheism Stance Corpus (ASC) \citep{wojatzki2016ltl}, comprising 715 tweets from the atheism portion of SemEval dataset. They demonstrated that word and sentence embeddings improved task performance, though not all variants performed equally. Comparing fastText, GloVe, and Universal Sentence Encoder (USE), they found USE achieved state-of-the-art results. However, they noted limited generalization assumptions given the corpus's focus on a single topic.  In contrast, we evaluated our approach on multiple corpora spanning various topics, providing a broader assessment of its generalizability across domains.
\citet{stede2020automatic} discussed the role of stance in argumentation mining, but not emotion or controversy. He focused more on theory, however, with some interesting examples which reflected how the task of finding a stance standpoint on a specific target topic can be complicated. He discussed how argumentation mining links to sentiment and stance detection—two popular tasks in computational linguistics often used as features in argumentation mining systems. He noted that while stance is closely related to argument, computing it accurately can be challenging. Meanwhile, straightforward sentiment systems relying only on pre-stored lexical polarities offer limited value and can often mislead. Therefore, we explicitly avoided using simple sentiment-based features, which can lead to misleading interpretations \citep[Section~7]{stede2020automatic}. Instead, we employed a context-aware emotion lexicon, designed to capture emotional nuances beyond simple lexical polarity. We demonstrated the initial potential of this flexible emotion lexicon for improving performance on the argumentative stance classification subtask.

\section{Development of eNRC: An Expansion of NRC}
\label{eNRC}

To improve emotion coverage in argumentative and stance-related text, we developed an approach to expand the NRC Emotion Lexicon. Although widely used, NRC misses many emotionally charged words common in modern language. We therefore propose an embedding-based expansion (eNRC) that groups semantically similar words and extends emotion labels while preserving categorical structure. The clustering step identifies semantic regions of emotional vocabulary, enabling us to normalize similarity scores and prevent category overlap. This ensures that the extended lexicon remains both comprehensive and interpretable.

Our experimental framework uses a batch-processing pipeline to compute DistilBERT embeddings and evaluate similarity scores. Cluster statistics (mean and standard deviation) for normalized similarity are precomputed. We use Hamming distance to measure the average difference between binary emotion vectors and entropy to quantify the diversity of emotion assignments. These metrics provide insights into the coherence and diversity of the expanded lexicon.


\subsection{Clustering and Embedding Distribution}
\begin{figure}[t]
\centering
\includegraphics[width=.45\textwidth]{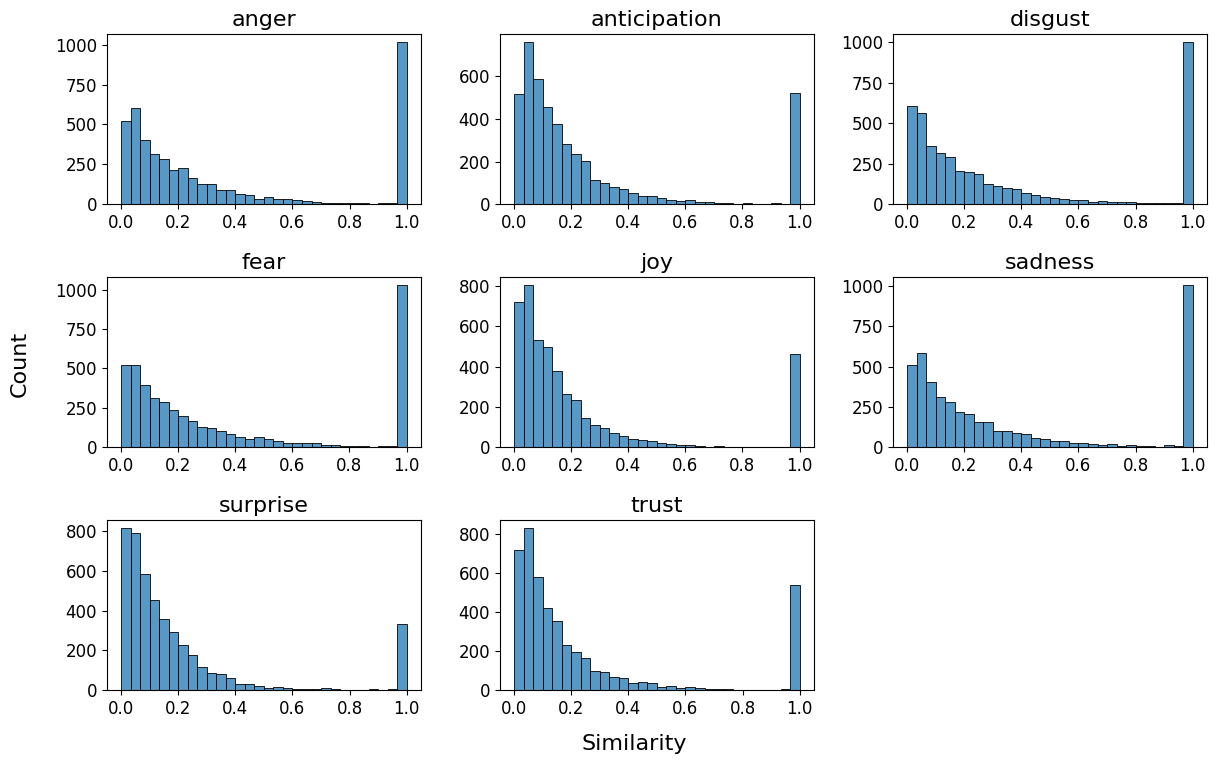}
\caption{Trimodal similarity distributions across emotion categories in the original NRC lexicon.}
\label{fig:nrc_vs_nrc}
\end{figure}

\paragraph{Trimodal Similarity Patterns}
We used DistilBERT to embed each word into a 768-dimensional space and to calculate cosine similarity. Figure~\ref{fig:nrc_vs_nrc} shows the similarity distribution between each NRC word embedding and all other words in the lexicon (including self-comparisons). The distributions are not uniform: each emotion exhibits a trimodal pattern, with the tall bar at a similarity of one corresponding to self-comparisons. This non-canonical shape suggests the presence of multiple density regions in the embedding space, which we explore next through dimensionality reduction and clustering.

\begin{figure}[t]
\centering
\includegraphics[width=.45\textwidth]{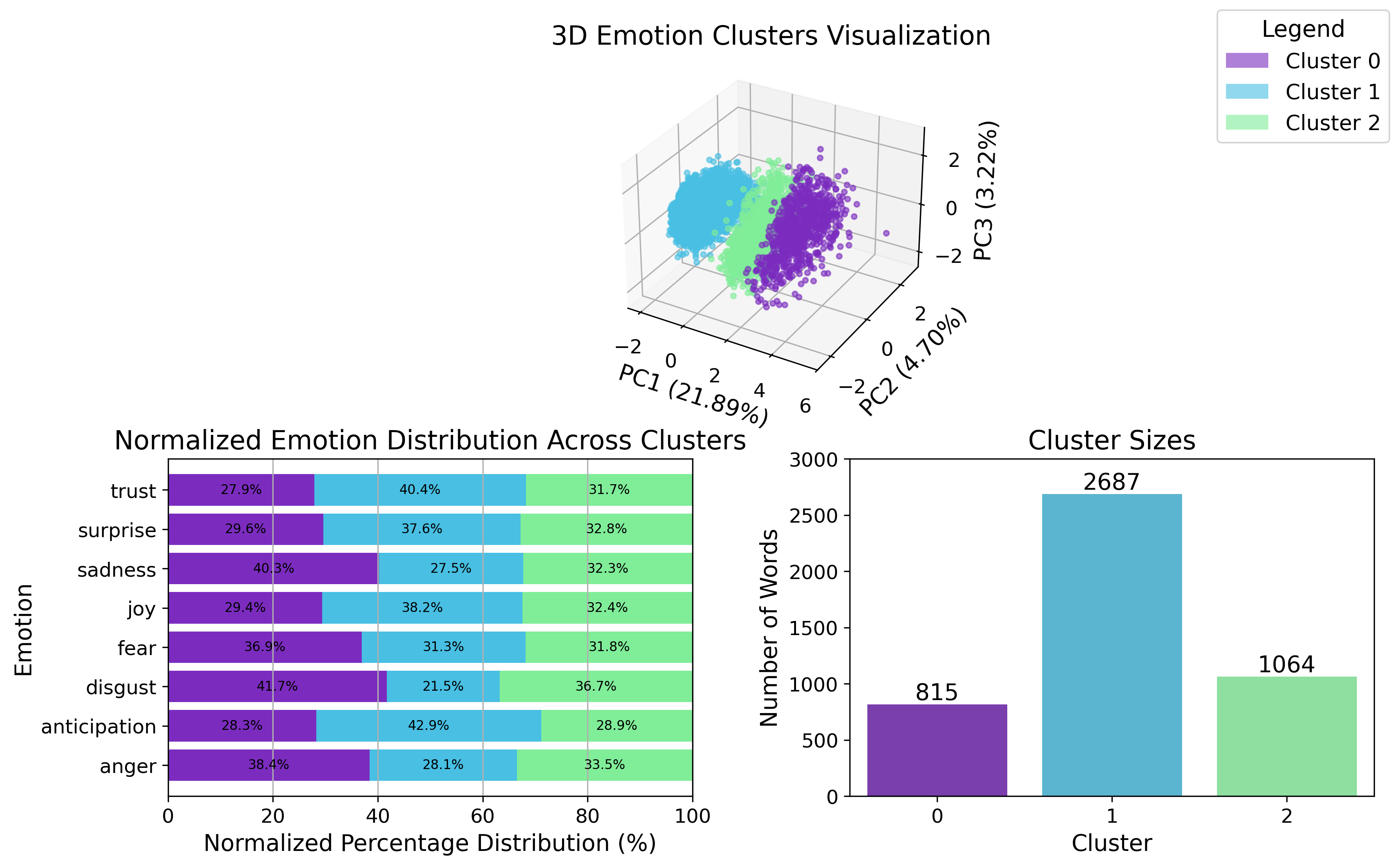}
\caption{Three-cluster structure of NRC word embeddings in PCA-reduced space}
\label{fig:cluster3d}
\end{figure}

\paragraph{Dimensionality Reduction and Clustering}
After observing a similar trimodal distribution in the similarity scores for each emotion, we reduced the embeddings to three dimensions using PCA \citep{hotelling1933analysis} to capture the dominant variance directions and facilitate clustering. Figure~\ref{fig:cluster3d} visualizes the embedding space after PCA reduction. We see three clusters that form well-separated, dense regions of semantically related NRC words. A Gaussian Mixture Model (GMM) \citep{dempster1977maximum} with three components partitions the space into clusters (violet, blue, and green), assigning each word to one of three dense regions representing distinct semantic distributions within the lexicon.

\paragraph{Cluster-based Normalization}

To account for varying density across clusters, we normalized similarity scores by first assigning each candidate and lexicon word to a cluster using PCA+GMM. We then computed similarity only among words within the same cluster—penalizing similarities between words from different clusters—and optionally rescale these scores to a standardized range within each cluster.

Our normalization follows the equation below. Let $s$ denote the raw similarity between two embeddings. For each cluster $i$, let $\mu_i$ and $\sigma_i$ be the precomputed mean and standard deviation of raw similarity scores, respectively. The normalized similarity for cluster $i$ is computed as
\[
s_{\text{norm}}^i = \text{clip}\left(\frac{\frac{s-\mu_i}{\sigma_i}+\sigma_c}{2\sigma_c},\,0,\,1\right),
\]
where $\sigma_c=3$ is a fixed scaling constant derived from the three-sigma rule. This mapping constrains values within $\pm3$ standard deviations of each cluster’s mean to the [0,1] range, clipping only extreme outliers. 

Finally, weighting by the cluster probabilities $p_i$, the overall normalized similarity is given by
\[
s_{\text{final}} = \text{clip}\left(\sum_{i=1}^{K} p_i\, s_{\text{norm}}^i,\,0,\,1\right),
\]
with $K=3$ being the number of clusters.

This normalization ensures that the similarity threshold used for lexicon expansion is applied consistently across different density regions.

\begin{figure}[t]
\centering
\includegraphics[width=.45\textwidth]{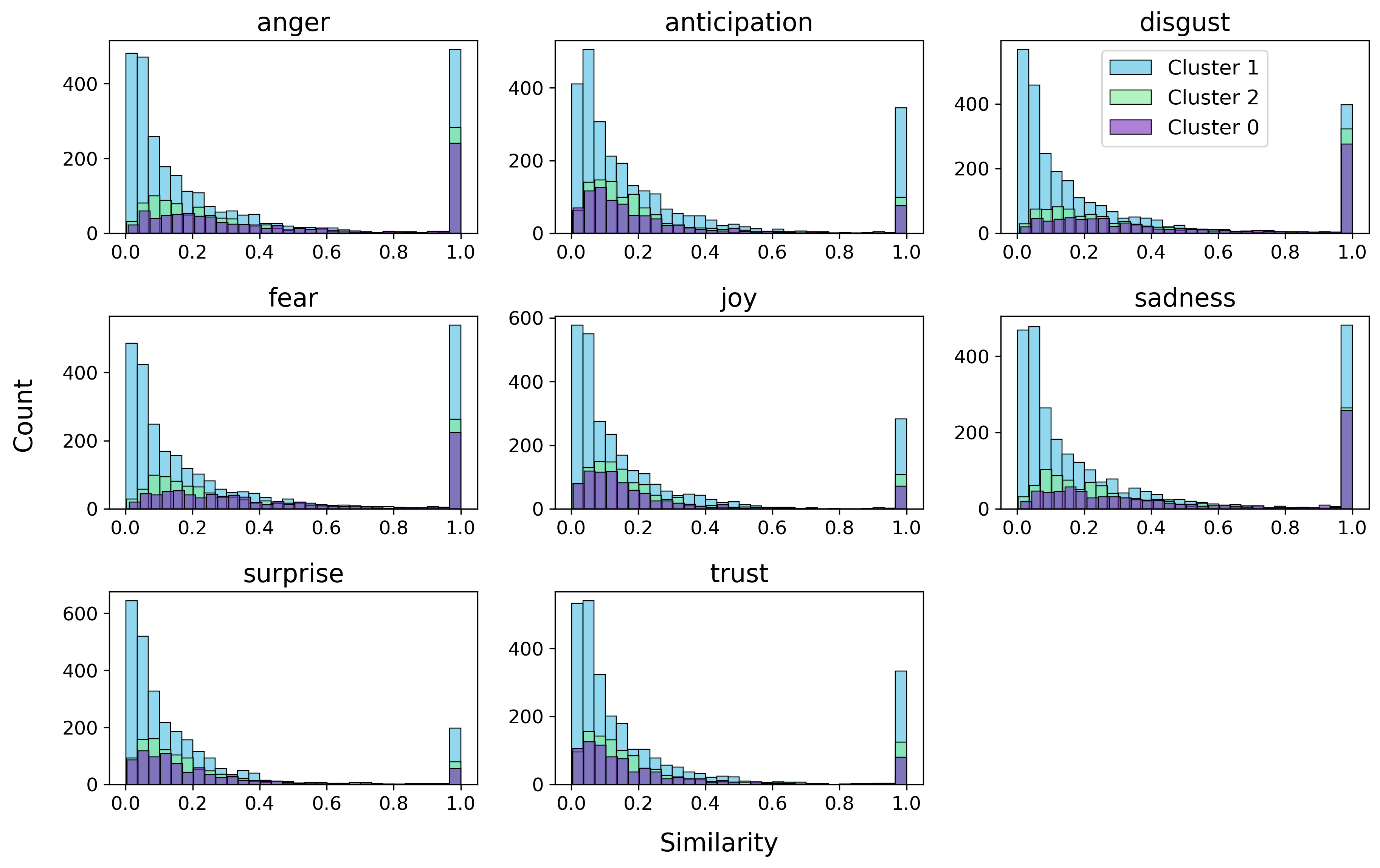}
\caption{Aligned similarity distributions after cluster-based normalization.}
\label{fig:emotion_histograms}
\end{figure}

Figure~\ref{fig:emotion_histograms} illustrates the similarity distributions after cluster normalization. The per-cluster curves now align in peak and shape, indicating that normalization harmonizes density differences across clusters and preserves emotion-category separability during expansion.

\subsection{Threshold-based Lexicon Expansion}
For each candidate word, we identified the nearest lexicon word for each emotion. If a calibrated similarity score exceeded a threshold $\theta$, we assigned the corresponding emotion to the candidate word. We varied $\theta$ from 0.05 to 0.95 in steps of 0.05 and recorded the number of new emotion assignments, the number of unique words expanded, diversity metrics (such as Hamming distance and entropy), and emotion-specific expansion counts.

\begin{figure}[t]
\centering
\includegraphics[width=.45\textwidth]{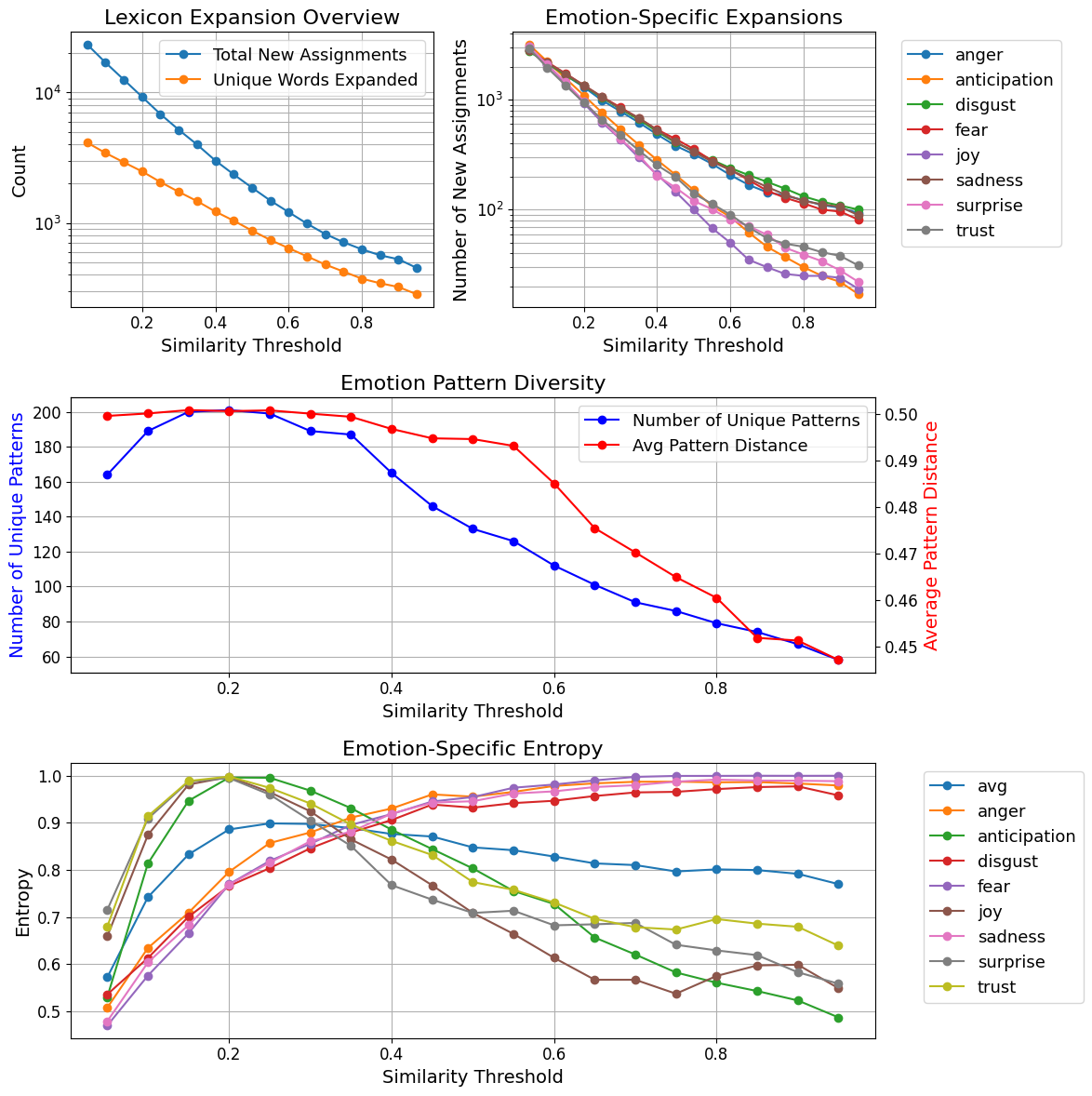}
\caption{Analysis of lexicon expansion under varying similarity thresholds.}
\label{fig:expansion-analysis}
\end{figure}

Figure~\ref{fig:expansion-analysis} shows the impact of varying the similarity threshold $\theta$ on lexicon expansion. As $\theta$ decreases, more emotion assignments occur, resulting in broader lexicon expansion. Crucially, however, lower thresholds risk introducing false positives by assigning emotions to words with only marginal similarity. Conversely, higher thresholds ensure more conservative expansion but might miss subtle emotion associations. The plots display total and unique new assignments (on a log scale), the number of unique emotion patterns, the average Hamming distance between binary emotion vectors, and trends in emotion-specific entropy. Importantly, the process is run on 9-dimensional feeling vectors (with 0 for absence and 1 for presence of an emotion), which change with each threshold.

\subsection{Threshold Selection and Category Separation}
The threshold calibration analysis revealed a clear trade-off between expansion coverage and structural coherence. Lower similarity thresholds led to rapid growth of the lexicon by assigning more emotion labels, but this also introduced semantic drift and weakened the separation between emotion categories. When the threshold drops below approximately 0.15, the extended lexicon reaches near-total corruption, with substantial cross-category contamination and loss of identifiable emotion boundaries. Higher thresholds, in contrast, restrict expansion and maintain strong category separation but limit lexical coverage. The optimal range lies where coverage is maximized without compromising the structural integrity of the emotion space. Within this range, emotion clusters remain distinct, preserving the categorical topology and ensuring that the expanded lexicon stays semantically consistent and empirically stable across clusters.



\section{Corpora and Statistics}
\subsection{Corpora}

\textbf{Argumentative Microtexts (Part 1)} The AMT1 corpus, developed by \citep{peldszus2015annotated}. Originally written in German, the texts have been professionally translated into English, as well as Russian \citep{fishcheva2019cross}, and more recently, Persian \citet{abkenar2024neural}. The corpus is annotated with complete argumentation tree structures.

\textbf{Argumentative Microtexts (Part 2)} The second part of the AMT2 corpus, developed by \citet{skeppstedt2018more} through crowdsourcing. It follows the same annotation approach as the original corpus, ensuring consistency. A notable difference in this corpus is the inclusion of implicit claims.

\begin{table*}[t]
\centering
\scriptsize
\setlength{\tabcolsep}{2.5pt} %
\renewcommand{\arraystretch}{1} 
\begin{tabular}{p{1.3cm}p{2cm}p{2.0cm}p{2.0cm}p{1.8cm}p{1.6cm}p{1.0cm}p{1.0cm}}
\toprule
\textbf{Corpora} & \textbf{Full Name} & \textbf{Domain} & \textbf{Size} & \textbf{Argument Types} & \textbf{Stance \newline Labels} & \textbf{\#Topics} & \textbf{Average \#Tokens}
\\
\midrule
\textbf{AMT1} & Microtext Corpus 1 & Short argumentative texts & 112  texts, 576 segments (ADUs) & Premise, Claim  & Pro/Opp & 19  & 14
\\
\textbf{AMT2} & Microtext Corpus 2 & Short argumentative texts &  116  texts, 614 segments (ADUs) & Premise, Claim & Pro/Opp & 34  & 13
\\
\textbf{PE} & Persuasive Essays & Student essays & 402  essays, 7,116 segments(ADUs), (1506 used) &  Premise, Claim, Major Claim & For/Against & 402 & 16
\\
\textbf{IBM} & IBM Debater Claim Stance & Debate arguments & 2,394 segments & Claim & Pro/Con & 55  & 12 \\
\textbf{UKP} & UKP Sentential \newline Argument Mining & Web discourse & 25,492 segments & Argumentative segment & For/Against/No & 8 & 24
\\
\bottomrule
\end{tabular}
\caption{Overview of original datasets before preprocessing and conversion. Actual subset sizes used in experiments are indicated where applicable.}
\label{tab:datasets_stat}
\end{table*}

\textit{Preprocessing Details:} In order to adapt the task for stance classification on the Microtext corpora, we projected the \textit{Pro} label in the XML files to \textit{For} and the \textit{Opp} label to \textit{Against}.  Since the number of samples was limited, we concatenated both corpora to create a combined dataset. 

\textbf{UKP} The UKP dataset comprises comments on various controversial topics \citep{stab2018cross}. Compared to other datasets, it contains the largest number of argumentative segments. The UKP comments are generally longer and often exhibit more complex argumentative structures, which increases the difficulty of stance classification. 

\textit{Preprocessing Details:}
To maintain consistency with other corpora, we focused on two labeling steps:  support or oppose segment, labeled as \textit{For} or \textit{Against}, and we ignore the \textit{no argument} category.

\textbf{Persusasive Essay} 
The PE corpus contains 402 argumentative essays written by English learners in response to specific prompts. Collected by \citep{Stab2016ParsingAS} from an online source, each essay is annotated with an argumentation graph. Essays begin with a guiding question and present a major claim, typically at the end, supported by evidence that may include sub-arguments. Some sentences serve a non-argumentative function, offering background or minor elaboration. 

\textit{Preprocessing Details:} The stance classification model is designed to predict the argumentative stance of each segment. The stance of a claim is specified through its stance attribute, as provided in the original corpus. We treat the major claim as the target topic and focus only on the stance of individual claims toward this target.

\textbf{IBM Argumentative Structure Dataset}
The IBM Argumentative Structure Dataset, originally introduced by  \citet{aharoni-etal-2014-benchmark}, includes claims about controversial topics. Each claim expresses support or opposition to its topic, and stance labels (\textit{Pro}/\textit{Con})  \citet{bar2017stance}.

\textit{Preprocessing Details:}
We preprocessed the IBM Debater dataset by extracting, for each debate, topics, claims, and stances, converting \textit{pro} and \textit{con} labels to \textit{For} and \textit{Against}.

\subsection{Corpora  Statistics}

\begin{figure}[t]
\centering
\includegraphics[width=.40\textwidth]{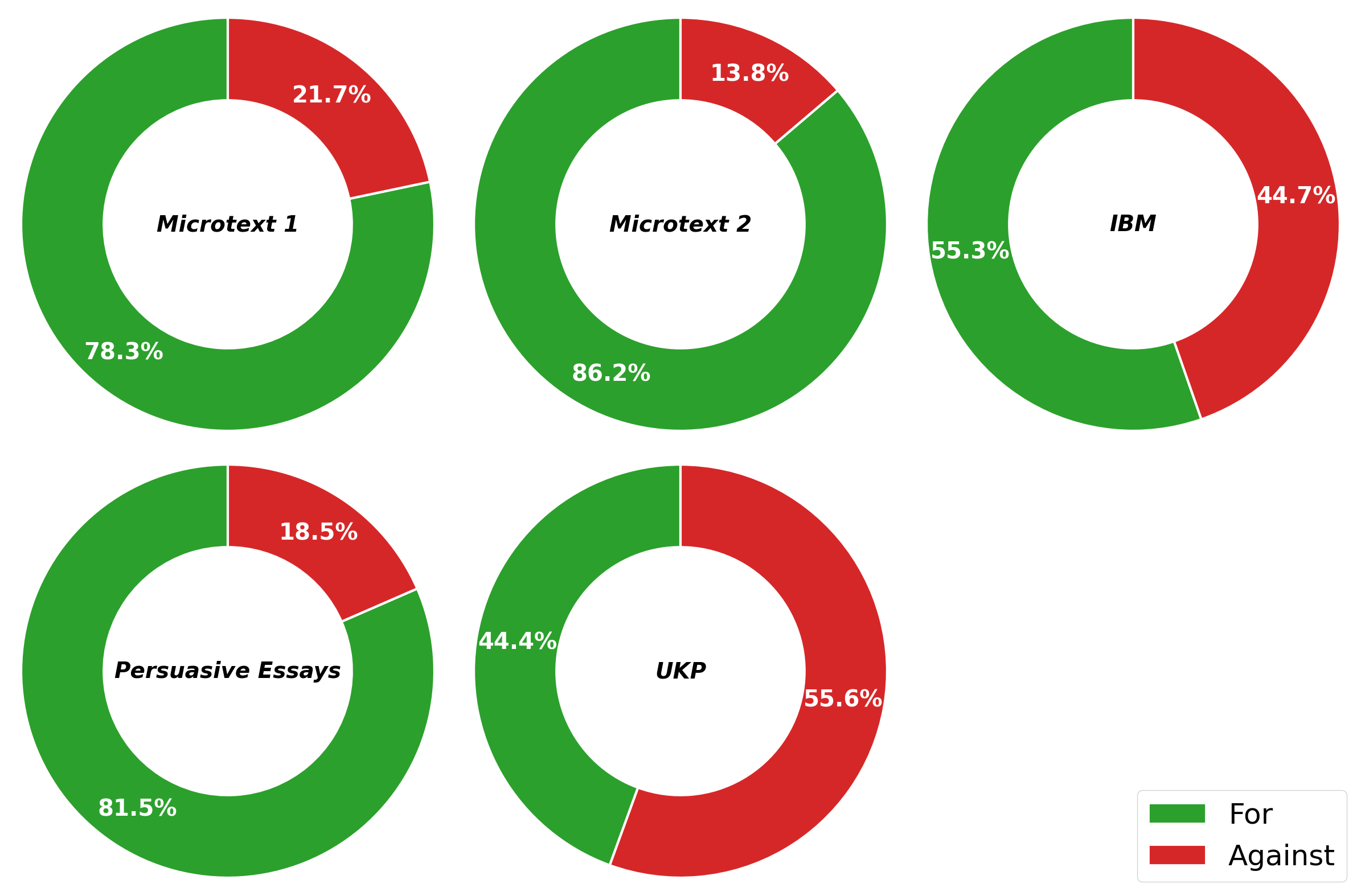}
\caption{Distribution of stance labels in the preprocessed corpora.}
\label{fig:am_corp_dist}
\end{figure}

Table~\ref{tab:datasets_stat} presents the statistics for each corpus, including domain, size and average number of tokens and Figure~\ref{fig:am_corp_dist} presents the distribution of their stance labels.

\section{Experiments and Discussion} 

\subsection{Model Architecture}

Let $\mathbf{x} = \{x_1, x_2, \ldots, x_n\}$ denote an input argumentative text segment consisting of $n$ tokens, and let $t$ represent the associated controversial topic. Our model aims to predict the stance $y \in \{0, 1\}$ (\textit{For} or \textit{Against}) of the text toward the topic.

We employ a pre-trained BERT encoder to obtain contextualized representations of the input. Following standard practice, we encode the text with explicit topic information by constructing the input as: ``[CLS] Topic: $t$ [SEP] Argument: $\mathbf{x}$ [SEP]''. This concatenation allows the model to jointly encode both the argumentative content and its target topic within the same semantic space.

The BERT encoder produces a sequence of hidden states $\mathbf{H} \in \mathbb{R}^{n \times d}$, where $d=768$ is the hidden dimension. We extract the [CLS] token representation $\mathbf{h}_{\text{CLS}} \in \mathbb{R}^d$ as the aggregate sentence embedding.

When additional emotion NRC features are available, we concatenate them with the [CLS] representation. Let $\mathbf{f}_{\text{emo}} \in \mathbb{R}^{d_e}$ denote the $d_e$-dimensional emotion feature vector derived from the extended NRC lexicon (eNRC). The combined representation $\mathbf{h}_{\text{combined}} = [\mathbf{h}_{\text{CLS}}; \mathbf{f}_{\text{emo}}] \in \mathbb{R}^{d+d_e}$ is then passed through the classifier for binary classification.

\begin{figure}[t]
\centering
\includegraphics[width=\columnwidth]{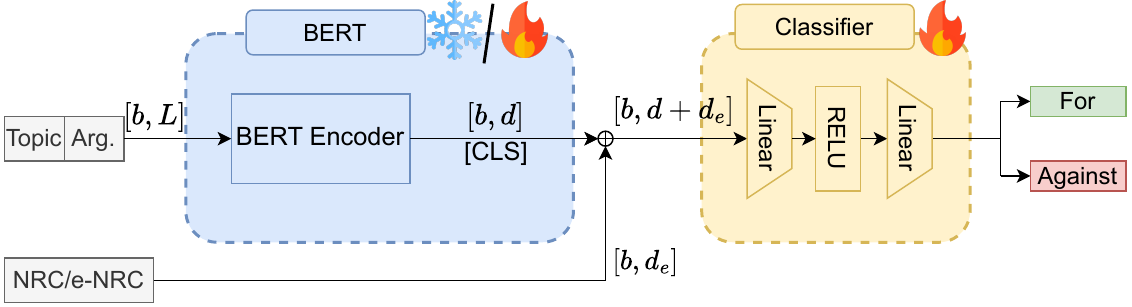}
\caption{Illustration of the NASCA model. The \textit{Topic} and \textit{Arg.} are composed into a single prompt and fed into the BERT model. Arg. stands for Argumentative segment.}
\label{fig:Model_arch_V2}
\end{figure}

\subsection{Experimental Setup}

We implement three variants of NASCA. \textbf{noNRC} refers to not using the extra NRC features. \textbf{NRC} refers to using the conventional NRC features. \textbf{eNRC} refers to using the expanded NRC features proposed by us. The classifier consists of two hidden layers. Each hidden layer applies ReLU activation followed by dropout regularization with rate $p=0.1$. The final output layer produces a single logit for binary classification. As baselines we choose majority class  (\textbf{MajC}) and a state-of-the-art LLM \textbf{Qwen2.5-7B}~\citep{qwen2025qwen25technicalreport}. For Qwen2.5-7B, we ask the model with the prompt: \textit{Please classify the following argument as ``support'' or ``agains'' to the given topic.} We split each dataset into train, test, and validation, and report the $F_1$ score on the test split using the best checkpoint on the validation split. 


\subsection{Results and Discussion}



Table~\ref{tab:main_table} presents the macro $F_1$ scores for stance classification across four datasets. The results demonstrate that incorporating emotion-lexicon features consistently improves performance, with eNRC achieving the best results across all datasets. Notably, both the majority class baseline (MajC) and the state-of-the-art LLM Qwen2.5-7B show limited effectiveness, with Qwen2.5-7B performing particularly poorly on UKP and PE, highlighting the challenge of stance classification in argumentative contexts even for large language models.

Our expanded emotion lexicon (eNRC) shows substantial improvements over both baseline approaches. The gains are particularly pronounced on the Persuasive Essays (PE) dataset, where eNRC achieves 62.9\% $F_1$, representing a 6.0 percentage point improvement over the noNRC baseline and a 2.6 percentage point gain over the original NRC. This significant boost reflects PE's longer essay format, which provides richer emotional content and more opportunities for the expanded lexicon to capture subtle emotional expressions. Similarly, the combined Microtext corpus (AMT1+2) shows strong improvement (+4.4 points over noNRC), suggesting that even in shorter argumentative texts, expanded emotion coverage enhances stance detection. The UKP dataset shows the highest absolute performance (68.4\% $F_1$ with eNRC), benefiting from both its larger size and the systematic emotion signal enhancement provided by the expanded lexicon.
The superior performance of eNRC over NRC stems from its ability to address fundamental coverage limitations in manually curated emotion lexicons. By leveraging DistilBERT embeddings and cluster-based normalization, our expansion method systematically identifies semantically similar words that carry similar emotional connotations. This approach proves particularly effective in the argumentation domain, where authors employ diverse emotional vocabulary to persuade and influence readers.

While our primary goal was not to surpass state-of-the-art (SOTA) results, but to ensure consistent performance across corpora and assess the effect of NRC lexicons on argumentative stance classification, our NASCA model with eNRC still achieved competitive results. As shown in Table \ref{tab:main_table}, it reached macro $F_1$ scores of 65.9\% on IBM (prev. SOTA: 64.5\% \citealp{bar2017stance}) and 68.4\% on UKP (prev. SOTA: 63.2\% \citealp{reimers-etal-2019-classification}). Results marked with {\dag} indicate cases where our model exceeded prior SOTA. For AMT1+2, no prior study evaluated both corpora jointly, and for PE, previous work used different task formulations.

\begin{table}[t]
\begin{center}
\begin{tabularx}{\columnwidth}{ccccc}
\toprule
      & \bf IBM            & \bf UKP            & \bf PE             & \bf AMT1+2            \\
\midrule
MajC  & 35.6          & 35.7          & 44.9          &  45.3    \\
Qwen2.5-7B &   36.2   & 30.7          & 38.9          & 62.3          \\
noNRC & 64.7          & 67.3          & 56.9          & 64.1         \\
NRC   & 64.8          & 65.4          & 60.3          & 63.4  \\
eNRC  & \textbf{65.9\textsuperscript{\dag}} & \textbf{68.4\textsuperscript{\dag}} & \textbf{62.9} & \textbf{68.5}       \\ 
\bottomrule
\end{tabularx}
\caption{Comparison of eNRC with other methods, majority class baseline (MajC), and Qwen2.5-7B on our corpora. We report the macro $F_1$ scores as the evaluation metric.}
\label{tab:main_table}
\end{center}
\end{table}

\begin{figure}[t]
\begin{center}
\includegraphics[width=.90\columnwidth]{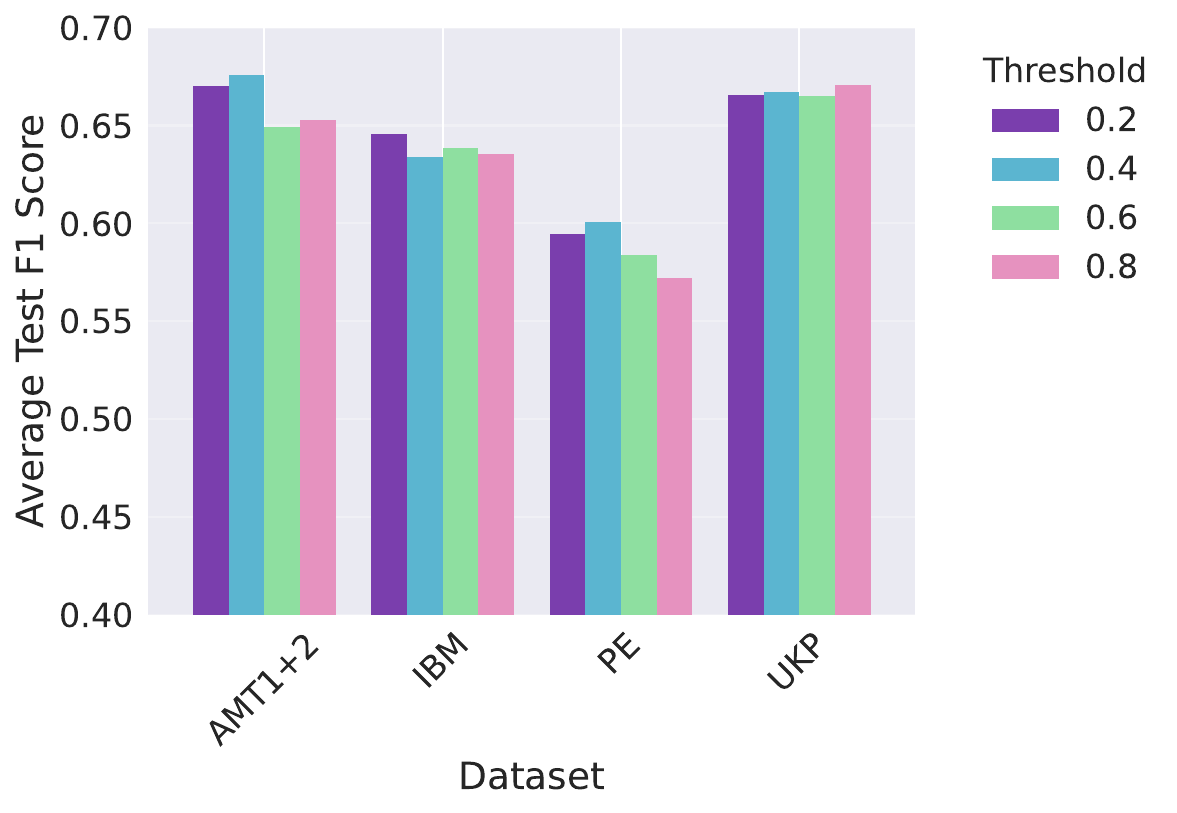}
\caption{Threshold analysis of eNRC. We report the macro $F_1$ scores using eNRC features with different thresholds.}
\label{fig:enrc}
\end{center}
\end{figure}

\paragraph{Sensitivity on eNRC Threshold.}
Figure \ref{fig:enrc}
reveals how the similarity threshold $\theta$ used during eNRC construction affects stance classification performance across all five datasets. This threshold determines the aggressiveness of lexicon expansion, with lower values assigning emotions to words with weaker similarity to NRC entries and higher values maintaining stricter semantic criteria.
The results demonstrate that $\theta = 0.4$ emerges as the most robust choice across datasets. This threshold achieves an effective balance between expanding emotion vocabulary coverage and maintaining semantic precision. 

Different datasets exhibit varying degrees of sensitivity to threshold selection, reflecting their distinct linguistic characteristics. The Persuasive Essays dataset shows pronounced sensitivity, with performance peaking sharply at $\theta = 0.4$ and declining more steeply at both extremes. This pattern aligns with PE's rich emotional language—the dataset requires expanded coverage to capture diverse emotional expressions, yet precision remains critical to avoid diluting the signal with false emotional associations. In contrast, IBM and UKP maintain relatively stable performance across a wider threshold range (0.2–0.8), suggesting these more formal argumentative texts rely on core emotional vocabulary that is already well-represented in the original NRC.

The Argumentative Microtexts datasets (AMT1+2) display moderate threshold sensitivity with optimal performance around $\theta = 0.4$, suggesting that the balanced expansion can effectively capture emotional signals in short texts where every word can carry substantial information.

\begin{figure}[t]
\begin{center}
\includegraphics[width=.90\columnwidth]{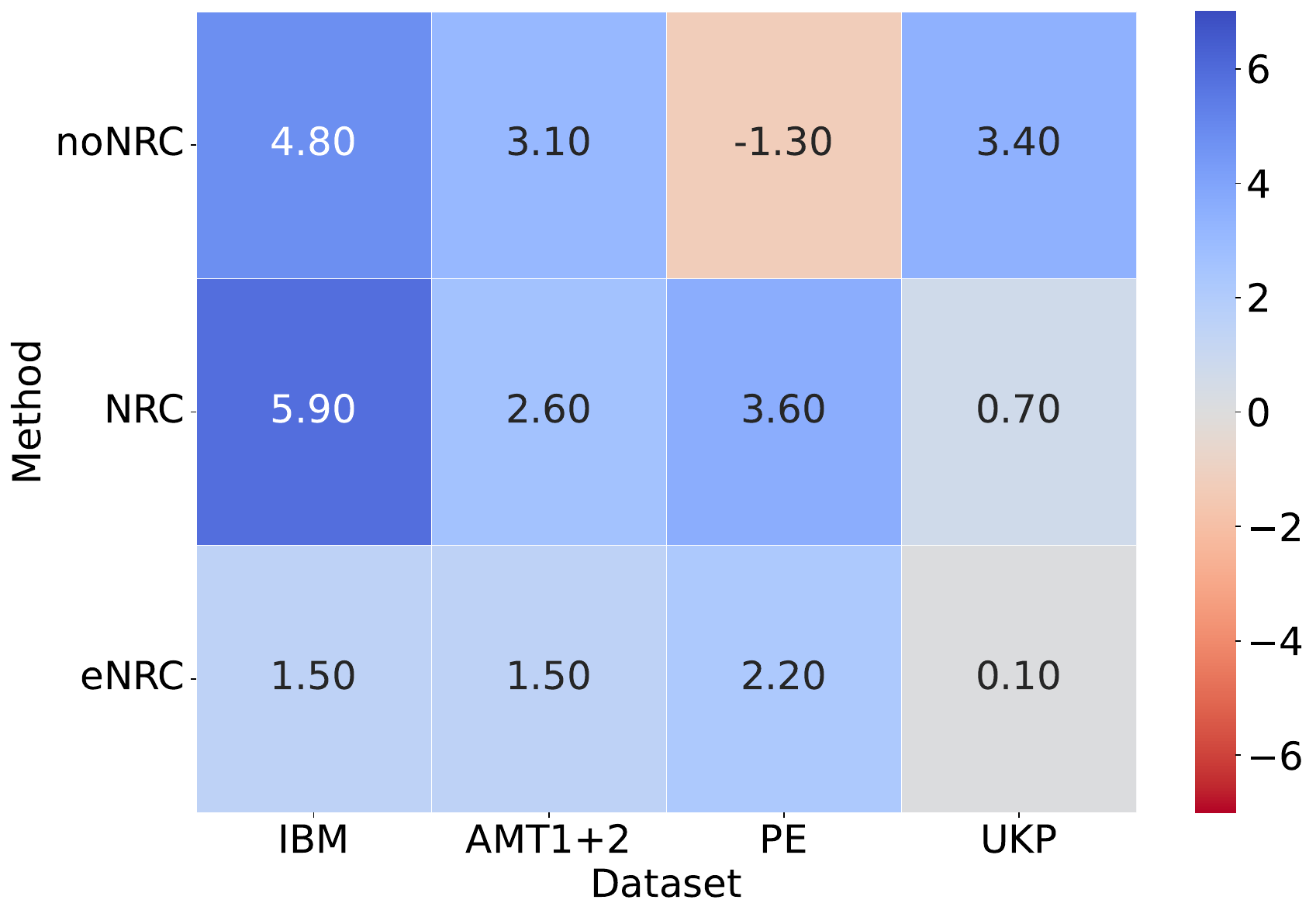}
\caption{Performance improvements after unfreezing the BERT model. We report the change in macro $F_1$ score.}
\label{fig:bert}
\end{center}
\end{figure}

\textbf{Impact of BERT Weights on the Performance.} Figure~\ref{fig:bert} presents the performance difference between models trained with frozen versus unfrozen BERT encoders across all datasets. While unfreezing BERT parameters typically yields performance gains in stance classification, our results reveal that this benefit diminishes substantially when incorporating eNRC features. For the eNRC variant, the performance differential remains within ±2.2 $F_1$ points across all datasets, compared to larger gaps observed with baseline approaches.

This reduced sensitivity to BERT fine-tuning suggests that the expanded emotion lexicon provides sufficiently discriminative features for stance classification, allowing the model to achieve competitive performance through classifier-level adaptation alone. The eNRC features appear to capture complementary information that partially compensates for the representational limitations of a frozen encoder. Notably, on the PE dataset, eNRC shows only +2.2 points improvement when unfreezing BERT, while noNRC and NRC show -1.3 and +3.6 points respectively, indicating that emotion features stabilize performance and reduce dependence on full model fine-tuning.

This finding has practical implications for deployment scenarios where computational resources are constrained, as it demonstrates that effective stance classification can be achieved with frozen BERT when augmented with high-quality emotion features. The ability to maintain performance without fine-tuning the entire encoder reduces training time, memory requirements, and computational costs while preserving model effectiveness.

\section{Conclusion}

In this article, we presented eNRC, an expansion of the NRC lexicon, integrated into an end-to-end approach for argumentative stance classification on controversial topics. Previous studies suffered from several limitations: they often focused on non-argumentative texts, were restricted to a single domain, or addressed only one corpus. In contrast, we reformulated five well-known corpora covering a range of controversial topics to evaluate our approach comprehensively.

To the best of our knowledge, this was the first study to systematically incorporate an emotion lexicon into this subtask, demonstrating its potential to improve performance. Additionally, we introduced this lexicon expansion through a rigorous and reproducible process, making it useful for other researchers. We also proposed a neural stance classification framework that could be easily adapted to other topics and domains with minimal modification. 

Both the expanded NRC lexicon and the argumentative stance classification framework are publicly available to the research community.

\clearpage

\nocite{*}
\section{Bibliographical References}\label{sec:reference}

\bibliographystyle{lrec2026-natbib}
\bibliography{paper}

\section{Language Resource References}
\label{lr:ref}
\bibliographystylelanguageresource{lrec2026-natbib}
\bibliographylanguageresource{languageresource}

\end{document}